%% file: main.tex
\documentclass[10pt,twocolumn,letterpaper]{article}

\usepackage{cvpr}              
\usepackage{colortbl}
\usepackage{graphics}
\input{preamble}

\definecolor{cvprblue}{rgb}{0.21,0.49,0.74}
\usepackage[pagebackref,breaklinks,colorlinks,citecolor=cvprblue]{hyperref}

\def\paperID{10527} 
\def\confName{CVPR}
\def\confYear{2024}

\title{Federated Active Learning for Target Domain Generalisation}

\author{Razvan Caramalau\\
University College London\\
London, UK\\
{\tt\small r.caramalau@ucl.ac.uk}
\and
Binod Bhattarai\\
University of Aberdeen\\
Aberdeen, UK \\
{\tt\small binod.bhattarai@abdn.ac.uk}
\and
Danail Stoyanov\\
University College London\\
London, UK\\
{\tt\small danail.stoyanov@ucl.ac.uk}
}

\begin{document}
\maketitle
\input{sec/0_abstract}    
\input{sec/1_intro}

\input{sec/2_related_works}
\input{sec/3_method}

\input{sec/4_experim}

\section{Conclusions and Future Work}
In this paper, we introduce FAL for DG for the first time, presenting FEDA, an FDG pipeline, and the associated AL selection function, FEDALV. With FEDA, we achieve state-of-the-art results on three FDG benchmarks, establishing a robust foundation in the generalisation process. To achieve this, we employ both local and global stages of optimisation. The local regularisation promotes simple representations and ensures alignment with a conditional distribution, while the global stage limits the free energy biases between the joint source domain and the target. Considering that client data is often non-iid, unavailable, or challenging to annotate, FEDALV simulates this FAL scenario on the same DG datasets. Empirical evidence and analyses demonstrate that our selection principle aligns newly labelled source samples with the target distribution. Moreover, our novel method outperforms both state-of-the-art in FAL (LoGo) and in active domain adaptation (EADA). For future work, we look forward to an extension analysis of FEDALV under a larger variation in clients and heterogeneity. Also, a more task-aware selection function could further enable the advantages of our proposal.

\section{Acknowledgements}
This work was supported in whole, or in part, by the Wellcome/EPSRC Centre for Interventional and Surgical Sciences (WEISS) [203145/Z/16/Z], the Department of Science, Innovation and Technology (DSIT) and the Royal Academy of Engineering under the Chair in Emerging Technologies programme; Horizon 2020 FET Open (EndoMapper) (12/2019 – 11/2024) [863146]. For the purpose of open access, the author has applied a CC BY public copyright licence to any author accepted manuscript version arising from this submission.
{
    \small
    \bibliographystyle{ieeenat_fullname}
    \bibliography{main}
}

\input{sec/X_suppl}

\end{document}

%% file: preamble.tex
\usepackage[dvipsnames]{xcolor}


%% file: sec/0_abstract.tex
\begin{abstract}

In this paper, we introduce Active Learning framework in Federated Learning for Target Domain Generalisation, harnessing the strength from both learning paradigms. Our framework, FEDALV, composed of Active Learning (AL) and Federated Domain Generalisation (FDG), enables generalisation of an image classification model trained from limited source domain client's data without sharing images to an unseen target domain. To this end, our FDG, FEDA, consists of two optimisation updates during training, one at the client and another at the server level. For the client, the introduced losses aim to reduce feature complexity and condition alignment, while in the server, the regularisation limits free energy biases between source and target obtained by the global model. The remaining component of FEDAL is AL with variable budgets, which queries the server to retrieve and sample the most informative local data for the targeted client. We performed multiple experiments on FDG w/ and w/o AL and compared with both conventional FDG baselines and Federated Active Learning (FAL) baselines. Our extensive quantitative experiments demonstrate the superiority of our method in accuracy and efficiency compared to the multiple contemporary methods. FEDALV manages to obtain the performance of the full training target accuracy while sampling as little as 5\% of the source client's data.
Please find the code at: \url{https://github.com/razvancaramalau/FEDALV.git}
\end{abstract}

%% file: sec/1_intro.tex
\section{Introduction}


Federated Learning (FL)~\cite{fed_challenges, fedavg, fedprox, fednoniid, scaffold, moon} is a burgeoning research area with applications across various domains, including computer vision~\cite{fedfacerecog, fedadas}, medical imaging~\cite{fed_melloddy, liu2021feddg}, and natural language processing~\cite{fedspeech}. FL addresses the challenge of training machine learning and deep learning models without sharing raw data, addressing concerns related to privacy protection.

The key advantage of FL lies in the diverse data distribution across local clients. However, in many scenarios, the overall data is not only heterogeneous but can also be unlabelled or corrupted. Heterogeneity in the data distribution among clients leads to a \emph{distribution or domain shift} problem. Early FL methods, such as FedAvg~\cite{fedavg}, did not account for this distribution shift between clients, resulting in poor generalization in such scenarios. Moreover, these FL methods often overlook the challenges related to data annotations, assuming that every client has overcome the problem of annotating data. In reality, data annotation is a laborious and costly task.

\label{sec:intro}

\begin{figure}
  \centering
 \includegraphics[trim=.5cm 1cm .5cm .0cm, clip, width=0.47\textwidth]{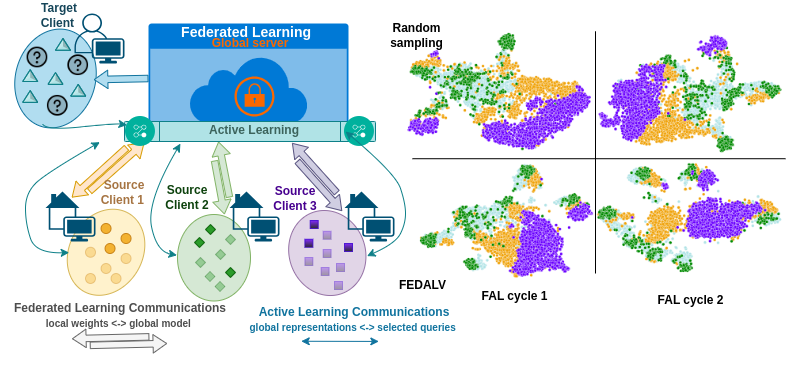}
  \caption{FEDALV High-level framework: Active Learning is integrated at the server level in our proposed FDG framework. Initially, a subset of source data is labelled, and over time, it is refined to retain only the most informative samples for target generalisation. Communication links with the global model operate during the FL training, and AL queries are propagated once the process is complete. After just a couple of cycles, the target domain (light blue) aligns with the sources in class clusters.}
  \label{fig:fedalv_intro}
\end{figure}

There is a growing research interest in FL to tackle the problem of domain generalisation~\cite{liu2021feddg,fedadg,flproto,fedsr,genadj_fldg,fdg_survey}, and these methods partly succeed in addressing the problem.
Generalising on the joint local domains can mitigate these shifts from the target domain. Therefore, standard aggregation techniques like FedAvg \cite{fedavg}, FedProx \cite{fedprox} and SCAFFOLD \cite{scaffold} can be further improved when accounting for the domain generalisation (DG) aspect. 
Unlike these methods, in this paper, we are introducing an Active Learning (AL)~\cite{dal,Nguyen2004ActiveLU} method to tackle the problem of domain generalisation in FL.
AL in FL for domain generalisation offers two-fold advantages. Firstly, it facilitates the identification of examples that closely resemble the target domain. Secondly, it avoids the annotation of redundant examples, addressing a key challenge in the learning process. AL has been successfully used to tackle the problem of domain generalisation~\cite{clue2021,eada}.
To the best of our knowledge, this is the first work to employ AL for domain generalisation in a federated manner. LoGo~\cite{logo} is one of the recent thorough studies of Federated Active Learning (FAL); however, it does not address the problem of domain generalisation.

In this paper, we present \textbf{FEDALV}, a novel framework of FAL for target DG. 
There are two important components: \textbf{FEDA}, (\textbf{F}ederated Learning pipeline with \textbf{E}nergy-based \textbf{D}omain \textbf{A}lignment) and an AL framework. FEDA induces domain-invariant representations locally and minimises the free energy alignment loss globally between the sources and targets in an unsupervised manner~\cite{eada}. On such representations and free energy biases, we employ
the AL framework to dynamically identify and query the most informative instances from the clients, directing the learning process towards regions critical for generalisation. On one hand, FEDA solves the FDG paradigm while on the other hand, FEDALV samples dynamically to limit data heterogeneity and possible future domain shifts. FEDALV accounts that some clients may contribute more or less to the generalisation problem.


As one of the critical demands of a FL pipeline, \emph{privacy} \cite{fed_challenges} must be assured between the clients. For example, leaking patient sensible information from a hospital would fringe several regulations \cite{liu2021feddg}.
Although in FEDA, the local optimisation stage imposes no risks, the server stage might raise some concerns. Therefore, we introduce a privacy policy after the communications round and aggregation of the global model. The policy guarantees that the global model is passed with updated parameters between each source and target data. Only the predictive classes are kept globally for the free energy alignment loss propagation. In this case no data is shared with the server and just the update cycle of the global model's parameters with its losses.




The contributions can be separated accordingly:
\begin{itemize}
    \item We introduce FEDA, a new FDG pipeline that proposes to regularise the representation complexity and the free energies generated between source and target clients.
    \item Following FEDA, we tackle for the first time the FDG with AL data selection. We design FEDALV criteria that can leverage informative unlabelled source samples to boost the performance on the unseen (zero-shot) domain.
    \item Through quantitative experiments and visual analysis, we demonstrate the efficacy of the introduced methods over three competitive benchmarks. In particular, we empirically evaluate them on three DG benchmarks: PACS \cite{pacs}, OfficeCaltech \cite{officecaltech} and OfficeHome \cite{officehome}. We show that FEDA achieves state-of-the-art results by a large margin when compared with recent baselines like FedSR or FPL. Moreover, FEDALV out-stands both classic AL sampling \cite{Sener2017ActiveApproach, ugcn} and more recent FAL strategies like LoGo \cite{logo}.
\end{itemize}




%% file: sec/2_related_works.tex
\section{Related Works}
\label{sec:rel_works}
Our proposed methodology of FEDA and FEDALV  stand at the congregation of three research fields: FL, DG and AL. Although it is the first work that combines all three, it is fair to discuss the advancements in their interactions: FDG, FAL and AL for DG.
\subsection{Federated Domain Generalisation}
FL has been intensively studied for handling the aggregation of distributed client models. The first and standard approach has been FedAvg \cite{fedavg}, where the local model parameters are averaged to obtain the global model. This has shown several issues with heterogeneous data and other optimisations of tracking gradient variance were introduced in FedProx \cite{fedprox} and SCAFFOLD \cite{scaffold}. However, these methods are still sub-optimal when the testing set is part of an unseen target domain.

Therefore, one of the first works in FDG is FedDG \cite{liu2021feddg}. Liu et al. propose an interpolation mechanism from shared frequency information between clients. Advancing this research, FedADG \cite{fedadg} adds a generative model locally as a reference to form aligned representations across domains. FedSR \cite{fedsr} follows a similar client-based strategy, but instead, it introduces two regularisers for aligning both the conditional and marginal representations. Due to its simple and effective form, we also add these losses in our proposed FDG, FEDA, for the local optimisation stage.

From another perspective, FDG has been approached by associating representations outside the client optimisation. Commonly, these methods use self-supervision to attract (FedDF \cite{feddf}) and repel inter-domain information. FCCL \cite{fccl} makes use of a logits cross-correlation matrix, while FPL \cite{flproto} combines two types of prototypes with their corresponding contrastive and regularisation losses. Differently from the mentioned methods, FedGA \cite{genadj_fldg} proposes a global-only optimisation where the aggregation weighting is adjusted according to prior performance.

\subsection{Federated Active Learning}
Although AL \cite{settles.tr09, dal} has been actively researched for deep learning \cite{Gal2017DeepDatab,Sener2017ActiveApproach,caramalau2021active,ugcn,Sinha2019VariationalLearning, ta-vaal,Yoo2019LearningLearning}, just recently some concepts were transferred in the context of FL. AFL \cite{ffal2019} was first introduced as a combined method to select both clients and unlabelled data. This, however, is applicable in limited scenarios. Thus, F-AL \cite{fal2022} considers annotators at each client and deploys classic AL selection methods like MCDropout \cite{Gal2016DropoutGhahramani}, CoreSet \cite{Sener2017ActiveApproach}, Learning Loss\cite{Yoo2019LearningLearning} and TA-VAAL \cite{ta-vaal}.

More dedicated FAL algorithms have just started to develop. KAFAL \cite{kafal2023} samples meaningful data by evaluating the knowledge discrepancies between clients and servers. Their methodology also adds its own FL update method. A more holistic analysis of FAL is investigated in LoGo \cite{logo} where diverse inter-class conditions were simulated (through local heterogeneity and global-imbalance ratio). Furthermore, LoGo proposes a robust selection function that uses metrics from both global and local models.
Despite all these FAL algorithms, FEDALV is the first work to sample source data for target generalisation. In the experiment section, we compare our selection with these prior works.

\subsection{Active Learning for Domain Generalisation}
AL in the context of DG has only been applied to adapt one domain to another by finding the informative target unlabelled samples. One of the first works, CLUE \cite{clue2021} combines uncertainty and clustering to query new data. The target domain selection is also explored in EADA \cite{eada} where it uses uncertainty and free-energy measurements for AL.

Despite this, none of these works are directly transferable and appropriate for the FDG pipeline due to structural and privacy reasons. Nevertheless, FEDALV exploits similar principles of free energy alignment present in EADA. Consequently, the client models are approximated as EBMs and the selection criteria tracks the free energy biases against the target samples. 

%% file: sec/3_method.tex
\section{Methodology}
\begin{figure*}[htp!]
  \centering
 \includegraphics[trim=0cm 0cm 0cm 0.0cm, clip, width=.9\textwidth]{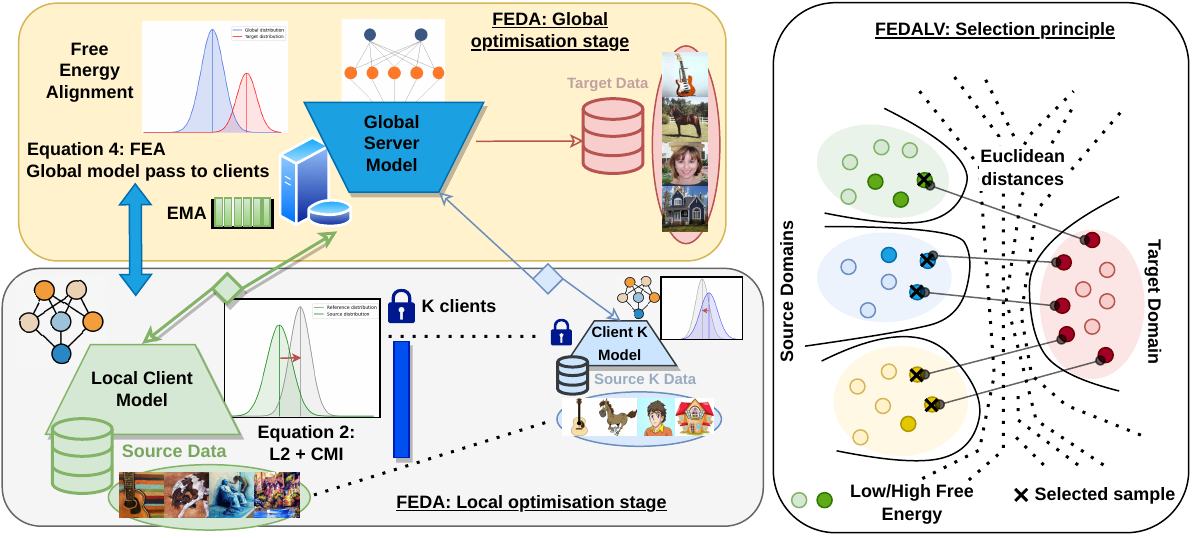}
  \caption{FEDA - Federated Learning with Energy-based Domain Alignment (\textbf{left}) is composed of two regularisation stages: the \textcolor{gray}{local} induces simple representations and alignment between sources; the \textcolor{orange}{global} exploits free energy biases between sources and target. 
  FEDALV Selection principle (\textbf{right}): ranks the target samples with the highest free energies and selects the source samples in their proximity.}
  \label{fig:feda}
\end{figure*}

As two of our main contributions are encompassed within the same framework, for simplicity, we divided the methodology into two parts. The first part introduces the components of FEDA (Fig. \ref{fig:feda} left), our FDG proposal. We commence with an overview of the FL pipeline and its aggregation technique. Subsequently, we delve into the two optimisation stages of FEDA that occur at the client and server levels. In the second part, we contextualize AL concepts within the realm of FDG through our method FEDALV. Finally, we provide a detailed account of its sampling strategy (Fig. \ref{fig:feda} right) in relation to FEDA.

\subsection{FEDA - Federated Learning with Energy-based Domain Alignment}

\subsubsection{Federated Learning setting}
The main objective of standard FL is to learn a global model $f_{g}(\theta_g): \mathcal{X}_g^K \rightarrow \mathcal{Y}$ from a set of collaborative $K$ clients. Here, for the classification task, $ \mathcal{X}_g^K $ is the joint input image space of all the clients, whilst $\mathcal{Y}$ are the expected fixed labels. Therefore, we can consider each client with its corresponding dataset $(x_k, y_k) \in \mathbf{D}_s^k$ and its local model function and parameters $f_{k}(\theta_k)$. We follow a classic FL training over rounds where the local model is trained and the global model $f_{g}(\theta_g)$ is updated at the communication round.

For FEDA, the communication round consists of local model parameters aggregation and the global model distribution back to the clients. While our aggregation is similar to FedAvg \cite{fedavg}, in the context of our application, we assign to each client a weighting according to its data space. Thus, the global model is formed at the communication round as:
\begin{equation}
    f_{g}(\theta_g) = \sum_{k=1}^K \mathbf{w}_k f_{k}(\theta_k),
    \label{eq:1}
\end{equation}
where $\mathbf{w}_k = N_{\mathbf{D}_s^k} / \sum N_{\mathbf{D}_s^k}$ ($N_{\mathbf{D}_s^k}$ is the dataset size for client $k$). 

Specifically concerning the DG task, the FL model also aims to perform well in unseen domains. For simplicity, let's define a single target client of the unseen domain 
$x_t \in \mathbf{D}_t$ where we assume similar labels $\mathcal{Y}$ as in the source domains. Standard DG techniques are not directly applicable to the FL setting mostly due to data-sharing in ensembles \cite{li2023simple_dg} or for fine-tuning \cite{mao2022contextaware_dg}.  However, in FEDA, we explore fundamental DG concepts, extracting domain-invariant representations and preserving alignment with the target domain while maintaining privacy. To achieve this, we present two optimization stages in the next sections.

\subsubsection{FEDA: Local model optimisation}
At each local round, the client (source) models are trained in parallel to optimise their parameters $\theta_K$.  To obtain domain-invariant representations, the models should produce representations with the same cross-domain distributions. Inspired by the work of FedSR \cite{fedsr}, we incorporate two regularisation losses aimed at reducing the local model's representation complexity and preserving essential information between inputs and representations.

An intuitive way of obtaining simple representation is by deploying an Euclidean norm regulariser over them. Given $z_k$ the latent representation of the client model $f_{k}(\theta_k)$, the L2-norm can be written as $\mathcal{L}^k_{L2} = 1 / N_{\mathbf{D}_s^k} \sum^{N_{\mathbf{D}_s^k}} || z_k ||_2^2$. If all clients follow a similar regularisation, there will be an implicit alignment to a normal reference distribution of the representations.

The second regularisation loss tracks the amount of mutual information from $x_k$ and $z_k$ by conditioning to the labels $y_k$. While $p_k(z_k | y_k)$ is not tractable, FedSR proposes an upper bound approximation with a conditional distribution $r_k(z_k | y_k)$. Here, $r_k$ can be a probabilistic representation network with classes as inputs. Thus, the outputs may follow a normal distribution. The conditional mutual information loss minimises the divergence between $p_k(z_k | x_k)$ and $r_k(z_k | y_k)$ accordingly: $\mathcal{L}^k_{CMI} = 1 / N_{\mathbf{D}_s^k} \sum^{N_{\mathbf{D}_s^k}} \mathbf{KL} [p_k(z_k | x_k) | r_k(z_k | y_k)] $. Linked to the conditional distribution alignment from DG, all $p_K(z_K | x_K)$ will aim to be closer to $r_k(z_k | y_k)$ 

Nevertheless, we advise the reader to look into the FedSR \cite{fedsr} paper for more proof and demonstrations of the $\mathcal{L}_{L2}$ and $\mathcal{L}_{CMI}$. By including the negative log-likelihood loss for classification, the local optimisation of FEDA for client $k$ is formulated as:
\begin{equation}
    \mathcal{L}^k_{client} = \mathcal{L}^k_{nll} +  \lambda_{L2} \mathcal{L}^k_{L2} + \lambda_{CMI} \mathcal{L}^k_{CMI},
    \label{eq:2}
\end{equation}
with $\lambda_{L2}$ and $\lambda_{CMI}$ scaling factors.

\subsubsection{FEDA: Global model optimisation}
Once the local models have been trained for a few rounds, the global model can be formed or updated during the communication round according to equation \ref{eq:1}. In the global optimisation, we view the new model as an energy-based model (EBM) \cite{ebms}. During training, the objective would be like finding an energy function $\mathcal{E}(x,y)$ that gives the lowest energies for the correct classes. 

In \cite{ebms,eada}, the probability of input $x$ is expressed as a free energy function $\mathcal{F}(x)$ "leaked" by an EBM: $p(x) = \frac{exp(-\mathcal{F}(x))}{\sum^{\mathcal{X}} exp(\mathcal{F}(x))}$. In this context, the joint probability of $p(x, y)$ is estimated through Gibbs distribution: $p(x,y) = exp(-\mathcal{E}(x,y))/ \sum^{\mathcal{X}}\sum^{\mathcal{Y}} exp(-\mathcal{E}(x,y))$. As $y \in \mathcal{Y}$ is fixed and $p(x) = \sum^{\mathcal{Y}} p(x,y)$, the free energy can be re-written as $\mathcal{F}(x)= - \log \sum^{\mathcal{Y}} exp(-\mathcal{E}(x,y))$.

If we consider all local models as EBMs and that FEDA aims to form a global EBM, the negative log-likelihood loss from equation \ref{eq:2} can be expressed at a client level as:

\begin{equation}
    \mathcal{L}^k_{nll}(x_k,y_k;\theta_k) = \mathcal{E}(x_k,y_k;\theta_k)) - \mathcal{F}(x_k;\theta_k),
    \label{eq:3}
\end{equation}
assuming that the reverse temperature is set to 1. The minimisation of this loss would create low  $\mathcal{E}$ for correct classifications, but the free energies will still be pulled up. As noticed in \cite{eada}, the free energies of the sources will still be lower than the ones from the target domain. Especially, this is the case in FEDA, where the local optimisation loop aligns all the source domains.

In global optimisation, we aim to reduce the free energy biases that are created between the source domains and the target. Assuming the local source alignment between clients, we introduce after aggregation another regularisation loss to minimise the effects of these biases. We denote it as free energy alignment loss:

\begin{equation}
    \mathcal{L}^k_{fea}(x_t;\theta_g) = \max(0, \mathcal{F}(x_t;\theta_g) - \mathrm{E}_{x_k \in \mathbf{D}_s^k} \mathcal{F}(x_k;\theta_g)).
    \label{eq:4}
\end{equation}

Due to privacy concerns, the first term of equation \ref{eq:3} is optimised at the client level while the $\mathcal{L}^k_{fea}(x_t;\theta_g)$ is trained in an unsupervised manner at the global optimisation. However, this regularisation requires free energy estimation for both source $x_k \in \mathbf{D}_s^k$ and target samples.

Therefore, we introduce a privacy policy for computing the $\mathcal{L}^k_{fea}$. The policy consists of a sequential parameter update $\theta^{'}_g$ where firstly $\mathcal{F}(x_t;\theta^{'}_g)$ is estimated from a batch of $\mathbf{x}_t \in \mathbf{D}_t$, followed by a set of source batches $\mathbf{x}_k$. The global model is passed between the target and the source client so that the gradient update with the regularisation happens in the server. The privacy control manages to share the global model between the clients and to store in the server the estimated energies of $\mathcal{F}(x_t;\theta_g)$ and $\mathcal{F}(x_k;\theta_g)$. It is worth mentioning that to obtain $\mathrm{E}_{x_k \in \mathbf{D}_s^k} \mathcal{F}(x_k;\theta_g)$ for each source client, similarly to \cite{eada}, we estimated it with the exponential moving average (EMA) over time. For more theoretical analysis on free energy, we refer \cite{eada} to the reader.  In relation to the global optimisation, we also scale the effect of the free energy alignment loss with $\lambda_{fea}$.


\subsection{FEDALV - Federated Active Learning with Energy-based Domain Alignment and local Variable budget}

In this section, we shift the objective of FDG with FEDA from an AL perspective. Thus, our main objective is to find the highest informative samples from the source clients that can help generalise to unseen target domains at a low labelling cost. We introduce FEDALV which follows a typical FAL pipeline but encapsulates FEDA's training strategy from the previous section.
First, we briefly present the structure of FAL after which we discuss the sampling technique. 

\subsubsection{Federated Active Learning setup}
Given a set of $K$ source clients and one $t$ target client/domain, we assume that each source has an initial pool of labelled data $\mathbf{D}_s^{L_0 (k)}$ and plenty of unlabelled data $\mathbf{D}_s^{U (k)}$. Distinctively from active domain adaptation (CLUE \cite{clue2021} and EADA \cite{eada}), instead of sampling informative from a target domain for adaptation, we propose to select new unlabelled samples from sources for DG. Therefore, similarly to LoGo \cite{logo}, the FAL framework demands available annotators (oracles) $\mathcal{O}_s^k$ at each client.

The exploration of the sample space in FAL is defined by the number of FL training cycles $C$, while the exploitation is limited by a budget $B_s^k$ of queries assigned to each source client. Fundamentally, the goal of our proposed FAL is to find the least number of unlabelled samples through a selection algorithm $\mathcal{A}(\mathbf{D}_s^{U (0)};\dots;\mathbf{D}_s^{U (K)})$ that can generalise to an unseen domain at a defined performance. From another perspective, our task can also be portrayed as a data curation application for FDG.

\subsubsection{FEDALV selection algorithm}
In AL, the selection algorithms are usually divided by their criteria into uncertainty-based, data diversity-based and a mixture of both.
As mentioned before, FEDALV sampling occurs after an FL training cycle of FEDA. Our proposed FDG pipeline, by default, intricately generates free energy biases between the source datasets and the target one. This can also be accounted as an uncertainty metric as these energies are pulled higher for the target than the source samples \cite{eada}. Therefore, to help the global regularisation term in FEDA, we propose to track the samples that have the highest free energies. Furthermore, the global model is used in the server for sampling as it is optimised to generalise well with FL.

At the $c \in C$ cycle, FEDALV initially evaluates the free energies $\mathcal{F}(x_t;\theta_g)$ from the unlabelled target space. Under the total budget $B = \sum^K B_s^k $, we select the samples with the highest energies and compute their representations $\mathbf{z}_t = f_g(\mathbf{x}_t;\theta_g)$. We also extract representations for the source clients and store everything at the server level for privacy concerns. Subsequently, by measuring the Euclidean distances between the samples, we select the closest source samples to the target one (with the highest free energy). Finally, the queried indices are sent back to the clients for labeling, initiating the next cycle of FEDA.

Although the overall budget $B$ is fixed, the number of queries selected from each client might differ. This is beneficial for the target domain as some domains may be less aligned to the others and so the global free energy biases can be reduced. Thus, the local client budget $B_s^k$ will vary from one AL cycle to another. In principle, due to finding source samples that have representation similarities with high-energy target images, the FEDALV selection principle is inherently uncertainty-based.

%% file: sec/4_experim.tex
\section{Experiments}
In the first part of our experiments, we evaluate the proposed FDG pipeline. FEDA is quantitatively compared with the current state-of-the-art on three datasets for image classification under a variation of model architecture and number of clients. For the second part, we deploy empirical quantification of FEDALV against other AL selection functions. Further analysis is investigated to strengthen our proposals.

\subsection{FEDA - Quantitative evaluation for Federated Domain Generalisation}
\subsubsection{Experiments setup}
\textbf{Datasets.} For benchmarking our domain generalisation performance, we leverage three well-known datasets, each with its own characteristics: PACS \cite{pacs}, OfficeHome \cite{officehome}, and OfficeCaltech \cite{officecaltech}. Each dataset contains data from four different domains: art, cartoon, photo, and sketch for PACS; art, clipart, product, and real for OfficeHome; and Caltech, Amazon, Webcam, and DSLR for OfficeCaltech. In terms of size and the number of categories, PACS contains 9,991 images over seven classes; OfficeHome has 15,500 images from 65 classes; and OfficeCaltech divides 2,533 images into ten classes. All of the benchmarks have heterogeneous data with respect to each domain.

\noindent \textbf{Federated Learning settings.} 
In validating FEDA, we conduct three sets of experiments. For PACS and OfficeHome, we adopt the FL settings from \cite{fedsr}, while for OfficeCaltech, we expand to more clients similarly to \cite{flproto}. Thus, FEDA trains three source clients over 100 rounds and has communication rounds with the server every five rounds on PACS and OfficeHome. On the other hand, for OfficeCaltech, we update three (Caltech), two (Amazon), one (Webcam), and four (DSLR) clients over the same number of rounds, but they communicate every ten rounds. Specifically for FEDA, we configure the three scaling factors for the regularisations accordingly: $\lambda_{fea} = 0.1$ on all datasets; PACS and OfficeCaltech - $\lambda_{L2} = 0.01$, $\lambda_{CMI} = 0.001$; and on OfficeHome - $\lambda_{L2} = 0.05$, $\lambda_{CMI} = 0.0005$. Moreover, we set a target batch size of 256 when computing the free energy alignment loss.

\noindent \textbf{Training settings.} 
We vary the client model architecture for each experiment depending on the complexity of the task: ResNet-10 \cite{he2016deep} on OfficeCaltech, ResNet-18 for PACS and the deeper ResNet-50 for OfficeHome. We noticed that keeping an SGD optimiser with a 0.01 learning rate, 1e-5 weight decay and 0.9 momentum has a stable training convergence for all models. Similarly to the other works \cite{fedsr,flproto,fccl}, the models are trained with a batch size of 64.

\noindent \textbf{Evaluation.} The evaluation of FEDA is done in a leave-one-out domain setting, thus, we alternate each domain as a target domain for our experiments. For each domain, we use 10\% of the data for validation and the rest for training (divided between clients) apart from the target. We measure the class accuracy of the global model and average it over 3 trials for each set of experiments.

\subsubsection{Comparison with state-of-the-arts}
\textbf{Baselines.} Our quantitative evaluations follow 2 different settings and for a fair comparison, we approach the state-of-the-art presented by following these benchmarks. Overall, we validate FedAvg \cite{fedavg} in these experiments, as the pioneer of FL aggregation. The baselines corresponding to the PACS and OfficeHome datasets (Tables \ref{tab:feda_pacs} \ref{tab:feda_officehome}) are FedADG \cite{fedadg}, FedSR \cite{fedsr} and FedGA \cite{fedg_genadj}. FedADG introduces an adversarial model at the client level to align all the models with a generated reference distribution and a class-wise regularisation. FedSR is our fundamental baseline for the local optimisation stage but with an even-weighted client aggregation.
FedGA, on the other hand, proposes a more generalisable way of aggregating to obtain the global model by taking into account the local performance prior to the communication round. This method is versatile and in \cite{fedg_genadj} it is deployed with other FDG techniques. To maintain a competitive baseline, we consider FedGA results as its best combination for PACS and OfficeHome datasets. While these baselines consider one domain per client, in the setting of OfficeCaltech (for more clients per domain) Table \ref{tab:feda_officecaltech}, we compare FEDA with MOON \cite{moon}, FedProto \cite{tan2022fedproto} and FPL\cite{flproto}. Both FedProto and FPL adopt prototypes of representations in different aggregation modes, however, only the latter is designed for the FDG problem. MOON makes use of standard contrastive learning losses \cite{simclr} to condition the local updates from the global model representations.

\begin{table}[htbp]
\centering
\resizebox{.4\textwidth}{!}{
\begin{tabular}{l|c|c|c|c|c}
\begin{tabular}[c]{@{}l@{}}FL Method / \\ Target Domain\end{tabular} & Art                      & Cartoon                  & Photo                    & Sketch                   & Average                  \\ \hline
FedAvg \cite{fedavg}                                & 77.8                     & 72.8                     & 91.9                     & 78.8                     & 80.3                     \\
FedADG \cite{fedadg}                                & 77.8                     & 74.7                     & 92.9                     & 79.5                     & 81.2                     \\
FedSR \cite{fedsr}                                  & 83.2                     & 76                       & \underline{93.8}                   & 81.9                     & 83.7                     \\
FedGA \cite{fedg_genadj}                           & 81.69                    & 77.23                    & 93.79                    & 82.75                    & 83.87                    \\
\cellcolor[HTML]{EFEFEF}\textbf{FEDA (ours)}                                                 & \cellcolor[HTML]{EFEFEF} \textbf{84.88} & \cellcolor[HTML]{EFEFEF} \textbf{84.04} & \cellcolor[HTML]{EFEFEF} 93.11 & \cellcolor[HTML]{EFEFEF} \textbf{84.23}& \cellcolor[HTML]{EFEFEF} \textbf{86.56}
\end{tabular}
}
 \caption{FEDA - \textbf{PACS} dataset - target performance [average over 3 trials]}
  \label{tab:feda_pacs}
\end{table}

\noindent \textbf{PACS} (Table \ref{tab:feda_pacs}). The ResNet-18 local models are initialised with ImageNet \cite{jia2009imagenet} weights before training, providing an advantage in classification performance for all the baselines. FEDA achieves top accuracy in 3 out of 4 target domains and yields the highest average of 86.56\% (+2.69 to FedGA). When compared to FedSR, which employs the same local regularisations, the importance of our global optimisation in FEDA is demonstrated by maintaining an accuracy of approximately 84\% over the Art, Cartoon, and Sketch domains.

\begin{table}[htbp]
\centering
\resizebox{.4\textwidth}{!}{
\begin{tabular}{l|c|c|c|c|c}
\begin{tabular}[c]{@{}l@{}}FL Method / \\ Target Domain\end{tabular} & Art                                    & Clipart                                & Product                                & Real                                   & Average                                \\ \hline
FedAvg \cite{fedavg}                                & 62.5                                   & 55.6                                   & 75.7                                   & 78.2                                   & 67.9                                   \\
FedADG \cite{fedadg}                                & 63.2                                   & 57                                     & 76                                     & 77.7                                   & 68.4                                   \\
FedSR \cite{fedsr}                                  & 65.4                                   & 57.4                                   & 76.2                                   & 78.3                                   &  69.3                             \\
FedGA \cite{fedg_genadj}                           & 58.14                                  & 54.44                                  & 73.76                                  & 75.74                                  & 65.53                                  \\
\cellcolor[HTML]{EFEFEF}\textbf{FEDA (ours)}                                                 & \cellcolor[HTML]{EFEFEF}\textbf{78.19} & \cellcolor[HTML]{EFEFEF}\textbf{62.02} & \cellcolor[HTML]{EFEFEF}\textbf{82.31} & \cellcolor[HTML]{EFEFEF}\textbf{83.83} & \cellcolor[HTML]{EFEFEF}\textbf{76.59}
\end{tabular}
}
 \caption{FEDA - \textbf{OfficeHome} dataset - target performance [average over 3 trials]}
  \label{tab:feda_officehome}
\end{table}

\noindent \textbf{OfficeHome} (Table \ref{tab:feda_officehome}). As in the previous experiment, the ResNet-50 starts with the same weight initialisation to address a more challenging classification task. In all target domain experiments, FEDA outperforms the other state-of-the-art methods in generalisation. The performance gap is even larger than 10\% for the Art domain. Therefore, the overall target average contributes a 7.29\% improvement to FedSR.

\begin{table}[htbp]
\centering
\resizebox{.4\textwidth}{!}{
\begin{tabular}{l|c|c|c|c|c}
\begin{tabular}[c]{@{}l@{}}FL Method / \\ Target Domain\end{tabular} & Caltech                       & Amazon                        & Webcam                        & DSLR                                   & Average                                \\ \hline
FedAvg \cite{fedavg}                                & 60.15                         & 75.44                         & 45.86                         & 36                                     & 54.36                                  \\
MOON \cite{moon}                                    & 56.19                         & 71.54                         & 41.04                         & 30.22                                  & 49.74                                  \\
FedProto \cite{tan2022fedproto}                                & \underline{64.03}                         & \underline{79.37}                         & 50.17                         & 40.33                                  & 58.47                                  \\
FPL \cite{flproto}                                 & 63.39                         & 79.26                         & \underline{55.86}                         & 48                                     & 61.63                                  \\
\cellcolor[HTML]{EFEFEF}\textbf{FEDA (ours)}                                                 & \cellcolor[HTML]{EFEFEF}57.92 & \cellcolor[HTML]{EFEFEF}74.13 & \cellcolor[HTML]{EFEFEF}50.95 & \cellcolor[HTML]{EFEFEF}\textbf{72.34} & \cellcolor[HTML]{EFEFEF}\textbf{63.84}
\end{tabular}
}
 \caption{FEDA - \textbf{OfficeCaltech} dataset - target performance [average over 3 trials]}
  \label{tab:feda_officecaltech}
\end{table}

\noindent \textbf{OfficeCaltech} (Table \ref{tab:feda_officecaltech}). To align with the settings of the \cite{flproto} experiments, the local models do not commence with pre-trained weights. As shown in Table \ref{tab:feda_officecaltech}, on average, FEDA generalises better than the new set of baselines like FedProto or FPL. The performance gain demonstrates that FEDA is robust to a diverse set of clients originating from different source domains. Moreover, it achieves the largest gap when there are 4 target clients in the DSLR domain.

\subsubsection{Ablation studies}
\begin{figure}[htbp]
  \centering
 \includegraphics[trim=0cm 0cm 0cm 0.0cm, clip, width=.41\textwidth]{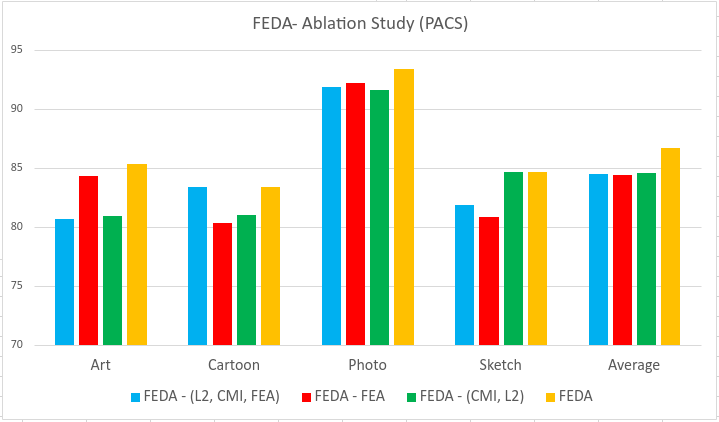}
  \caption{FEDA: Ablation evaluation [Zoom in for closer values]}
  \label{fig:feda_abla}
\end{figure}
We further justify the relevance and impact of the local and global optimisation stages by empirically evaluating FEDA on the PACS dataset. Figure \ref{fig:feda_abla} presents the chart of target performance from our method with no regularisation (FEDA - L2, CMI, FEA in blue) to the full version (in yellow). It is clearly observable that the local regularisations (of L2 and CMI) favour some target domains, while the global-only stage version outperforms others (Cartoon and Sketch). When averaging, the three variants have a close accuracy with a slight gain for the global-only FEDA. Nevertheless, the collaborative nature between the local and global stages maintains a consistent performance of our method on all target clients.

\subsection{FEDALV Quantitative evaluation with other Active Learning selection functions}
\subsubsection{Experiments setup}
As displayed in Tables \ref{tab:fedalv_pacs} \ref{tab:fedalv_officehome} \ref{tab:fedalv_officecaltech},  we train FEDA over $C=5$ AL cycles. For each experiment, we randomly sample and request annotations for an initial set of either 1\% (for PACS and OfficeHome) or 10\% (for OfficeCaltech) of the client's available data. We also define the budget $B$ of the same size from all the unlabelled source data. In terms of training, we optimise the models with the settings from FEDA. However, due to smaller batches, the performance flattens after 30-50 epochs. Furthermore, to ease the evaluation on OfficeCaltech, we keep the one client per domain setting as in the other experiments and upgrade the learner to ResNet-18.
To evaluate the generalisation robustness, we present the average of the target clients over three FAL trials.
\subsubsection{Comparison with other selection functions}
\textbf{Baselines.} As a standard and naive way of selecting data, local random uniform sampling is considered to compare with AL methods. Drawing insights from the most comprehensive study of sampling in FAL, we reference LoGo \cite{logo}, which combines uncertainty and clustering at both global and local levels and has proven to be the most effective in its set of experiments. The second competitive baseline in their results is CoreSet \cite{Sener2017ActiveApproach}. This method minimises the distances between the representations of labelled samples and a core set of unlabelled images. Lastly, EADA \cite{eada} is deployed as the state-of-the-art for active domain adaptation. The selection principle is similar to FEDALV as it relies on free energy measurement between target and source; however, EADA focuses on sampling from the target and includes class uncertainty.

All of these baselines are sampled with a fixed client budget as a percentage of available unlabelled data. In contrast, FEDALV allows the exploitation to affect the clients with variation. Thus, we include in our comparisons FEDAL, which maintains the same selection but under a constrained local budget.

\begin{table}[htbp]
\centering
\resizebox{.42\textwidth}{!}{
\begin{tabular}{l|c|c|c|c|c}
\begin{tabular}[c]{@{}l@{}}FAL Method / \\ \% of labelled data\end{tabular} & 1\%                           & 2\%                                    & 3\%                                    & 4\%                                    & 5\%                                    \\ \hline
Random                                                                      & \cellcolor[HTML]{FFFFFF}56.21 & \cellcolor[HTML]{FFFFFF}68.16          & \cellcolor[HTML]{FFFFFF}71.28          & \cellcolor[HTML]{FFFFFF}75.38          & \cellcolor[HTML]{FFFFFF}76.5           \\
CoreSet \cite{Sener2017ActiveApproach}                                            & \cellcolor[HTML]{FFFFFF}55.66 & \cellcolor[HTML]{FFFFFF}66.21          & \cellcolor[HTML]{FFFFFF}70.65          & \cellcolor[HTML]{FFFFFF}75.38          & \cellcolor[HTML]{FFFFFF}77.43          \\
LoGo \cite{logo}                                           & \cellcolor[HTML]{FFFFFF}55.34 & \cellcolor[HTML]{FFFFFF}64.58          & \cellcolor[HTML]{FFFFFF}69.61          & \cellcolor[HTML]{FFFFFF}70.99          & \cellcolor[HTML]{FFFFFF}72.46          \\
EADA \cite{eada}                                           & \cellcolor[HTML]{FFFFFF}55.29 & \cellcolor[HTML]{FFFFFF}67.05          & \cellcolor[HTML]{FFFFFF}73.94          & \cellcolor[HTML]{FFFFFF}76.93          & \cellcolor[HTML]{FFFFFF}79.99          \\
FEDAL (ours)                                               & \cellcolor[HTML]{FFFFFF}54.39 & \cellcolor[HTML]{FFFFFF}66.42 & \cellcolor[HTML]{FFFFFF}68.58 & \cellcolor[HTML]{FFFFFF}72.24 & \cellcolor[HTML]{FFFFFF}72.95 \\
\cellcolor[HTML]{FFF2CC}\textbf{FEDALV (ours)}                                                     & \cellcolor[HTML]{FFF2CC}55.24 & \cellcolor[HTML]{FFF2CC}\textbf{71.72} & \cellcolor[HTML]{FFF2CC}\textbf{78.07} & \cellcolor[HTML]{FFF2CC}\textbf{81.47} & \cellcolor[HTML]{FFF2CC}\textbf{83.63}
\end{tabular}
}
 \caption{FEDALV - \textbf{PACS} dataset - AL target performance [average over 3 trials]}
  \label{tab:fedalv_pacs}
\end{table}

\begin{table}[htbp]
\centering
\resizebox{.42\textwidth}{!}{
\begin{tabular}{l|c|c|c|c|c}
\begin{tabular}[c]{@{}l@{}}FAL Method / \\ \% of labelled data\end{tabular} & \multicolumn{1}{c|}{1\%}      & \multicolumn{1}{c|}{2\%}               & \multicolumn{1}{c|}{3\%}               & \multicolumn{1}{c|}{4\%}               & \multicolumn{1}{c}{5\%}                \\ \hline
Random                                                                      & \cellcolor[HTML]{FFFFFF}24.08 & \cellcolor[HTML]{FFFFFF}32.39          & \cellcolor[HTML]{FFFFFF}36.77          & \cellcolor[HTML]{FFFFFF}42.25          & \cellcolor[HTML]{FFFFFF}43.68          \\
CoreSet \cite{Sener2017ActiveApproach}                                            & \cellcolor[HTML]{FFFFFF}24.76 & \cellcolor[HTML]{FFFFFF} \underline{33.6}           & \cellcolor[HTML]{FFFFFF}40.06          & \cellcolor[HTML]{FFFFFF}45.02          & \cellcolor[HTML]{FFFFFF}47.11          \\
LoGo \cite{logo}                                           & \cellcolor[HTML]{FFFFFF}24.02 & \cellcolor[HTML]{FFFFFF}31.53          & \cellcolor[HTML]{FFFFFF}35.25          & \cellcolor[HTML]{FFFFFF}40.57          & \cellcolor[HTML]{FFFFFF}43.08          \\
EADA \cite{eada}                                           & \cellcolor[HTML]{FFFFFF}24.06 & \cellcolor[HTML]{FFFFFF} 33.28          & \cellcolor[HTML]{FFFFFF}36.28          & \cellcolor[HTML]{FFFFFF}42             & \cellcolor[HTML]{FFFFFF}45.48          \\
FEDAL (ours)                                               & \cellcolor[HTML]{FFFFFF}24.73 & \cellcolor[HTML]{FFFFFF}30.28 & \cellcolor[HTML]{FFFFFF}35.58 & \cellcolor[HTML]{FFFFFF}39.74 & \cellcolor[HTML]{FFFFFF}42.89 \\
\cellcolor[HTML]{FFF2CC}\textbf{FEDALV (ours)}                                                     & \cellcolor[HTML]{FFF2CC}24.3  & \cellcolor[HTML]{FFF2CC} 32.92 & \cellcolor[HTML]{FFF2CC}\textbf{41.49} & \cellcolor[HTML]{FFF2CC}\textbf{46.57} & \cellcolor[HTML]{FFF2CC}\textbf{50.04}
\end{tabular}
}
 \caption{FEDALV - \textbf{OfficeHome} dataset - AL target performance [average over 3 trials]}
  \label{tab:fedalv_officehome}
\end{table}

\begin{table}[htbp]
\centering
\resizebox{.42\textwidth}{!}{
\begin{tabular}{l|c|c|c|c|c}
\begin{tabular}[c]{@{}l@{}}FAL Method / \\ \% of labelled data\end{tabular} & \multicolumn{1}{c|}{10\%}     & \multicolumn{1}{c|}{20\%}     & \multicolumn{1}{c|}{30\%}     & \multicolumn{1}{c|}{40\%}     & \multicolumn{1}{c}{50\%}      \\ \hline
Random                                                                      & \cellcolor[HTML]{FFFFFF}32.17 & \cellcolor[HTML]{FFFFFF}43.6  & \cellcolor[HTML]{FFFFFF}42.38 & \cellcolor[HTML]{FFFFFF}57.29 & \cellcolor[HTML]{FFFFFF}62.71 \\
CoreSet \cite{Sener2017ActiveApproach}                                            & \cellcolor[HTML]{FFFFFF}34.77 & \cellcolor[HTML]{FFFFFF}43.42 & \cellcolor[HTML]{FFFFFF}50.13 & \cellcolor[HTML]{FFFFFF}59.97 & \cellcolor[HTML]{FFFFFF}66.12 \\
LoGo \cite{logo}                                           & \cellcolor[HTML]{FFFFFF}32.95 & \cellcolor[HTML]{FFFFFF}48.36 & \cellcolor[HTML]{FFFFFF}47.2  & \cellcolor[HTML]{FFFFFF}60.08 & \cellcolor[HTML]{FFFFFF}63.81 \\
EADA \cite{eada}                                           & \cellcolor[HTML]{FFFFFF}36.76 & \cellcolor[HTML]{FFFFFF}45.46 & \cellcolor[HTML]{FFFFFF}36.68 & \cellcolor[HTML]{FFFFFF}63.25 & \cellcolor[HTML]{FFFFFF}68.28 \\
FEDAL (ours)                                               & \cellcolor[HTML]{FFFFFF}34.89 & \cellcolor[HTML]{FFFFFF}50.22 & \cellcolor[HTML]{FFFFFF}53.66 & \cellcolor[HTML]{FFFFFF}65.04 & \cellcolor[HTML]{FFFFFF}70.45 \\
\rowcolor[HTML]{FFF2CC} 
\cellcolor[HTML]{FFF2CC}\textbf{FEDALV (ours)}                             & 33.09                         & \textbf{59.93}                & \textbf{66.37}                & \textbf{68.51}                & \textbf{72.34}               
\end{tabular}
}
 \caption{FEDALV - \textbf{OfficeCaltech} dataset - AL target performance [average over 3 trials]}
  \label{tab:fedalv_officecaltech}
\end{table}

\begin{figure*}[!h]
  \centering
 \includegraphics[trim=1cm .2cm 1cm 0.2cm, clip, width=.85\textwidth]{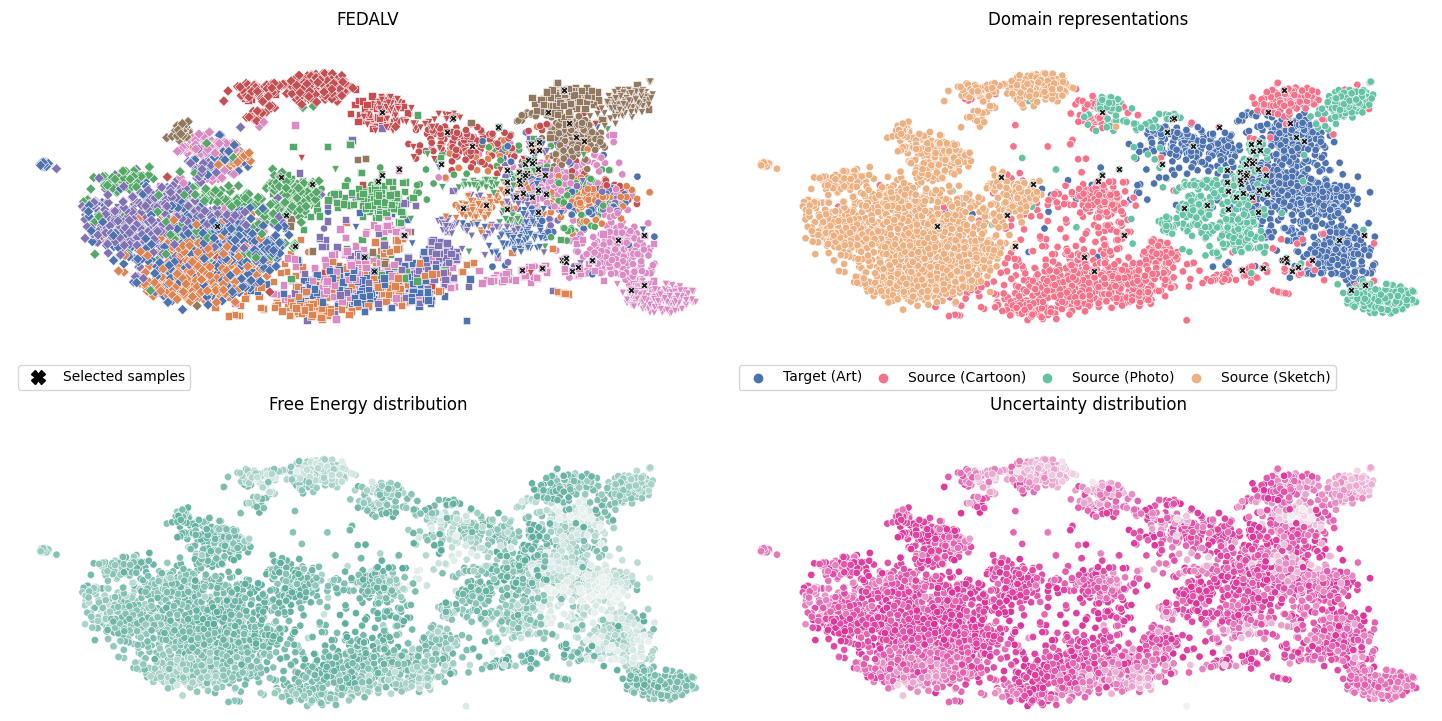}
  \caption{FEDALV: Selection Analysis [Zoom in for a better view]}
  \label{fig:fedalv_qual}
\end{figure*}
Quantitatively, FEDALV achieves state-of-the-art performance on all three FAL benchmarks, as presented in Tables \ref{tab:fedalv_pacs}, \ref{tab:fedalv_officehome}, and \ref{tab:fedalv_officecaltech}. Our method can sample representative source samples to such an extent that, with just 5\% of labelled data, it matches the FedSR performance on the entire PACS dataset (83.63\%). As expected, other AL selection techniques prove to be sub-optimal in the FDG context. LoGo, despite being the state-of-the-art for FAL, performs as well as random sampling. On PACS and OfficeCaltech, EADA shows competitive results, demonstrating the benefits of measuring free energy biases. When using even local budgets, FEDAL yields significant results, particularly in OfficeCaltech where the selection is higher (10\%). This is attributed to the lack of diversity that some source domains may have for domain generalisation.

\subsection{FEDALV Selection Analysis}

Prior to deciding on the FAL selection for DG, we conducted several analyses, and some are presented in Fig. \ref{fig:fedalv_qual}. This includes a t-SNE \cite{Maaten08visualizingdata} visualisation of the global model's latent features calculated for PACS. The data is unlabelled and from all domains, and the features are extracted after the first FAL cycle. In the top-left side, the classes are shown in distinct colours, and the domains are represented with different shapes. A clearer domain distribution is in the top-right part, where the target domain (Art) collides with the Photo domain and a small portion of the Cartoon source. In the bottom-left, we have the value distribution of the free energies for the same unlabelled samples, confirming that the highest values (displayed brighter) correspond to the target client. Gathering these insights, FEDALV marks with crosses the unlabelled source images at the border of the Art domain, indicating a preference for selecting more samples from the Photo domain. Moreover, our selection principle also identifies samples close to the target ones that are far from the cluster, occurring right at the borders between the Sketch and Cartoon domains.

Lastly, in the bottom-right representation, class uncertainty is measured from the global model (brighter is higher). FEDALV is not entirely dependent on this measurement, but it can support the classifier when the selected samples are close to the target's uncertain areas.

%% file: sec/X_suppl.tex
\clearpage
\setcounter{page}{1}
\maketitlesupplementary
\section{FEDA - relugarisation scalers tuning}
Our proposed solution for FDG introduces two local regularisations $\mathcal{L}_{L2}$ and $\mathcal{L}_{CMI}$ and one global optimisation loss $\mathcal{L}_{fea}$. These regularisations are attenuated by their corresponding scaling factors $\lambda_{L2}$, $\lambda_{CMI}$, and $\lambda_{fea}$. When increasing the local scalers, we noticed a smoothness in the local representations from the first local rounds, indicating a lower source client performance. In this way, both the local and global accuracies (after communication round) are affected by this degradation when the models are not learning more complex domain-invariant representations. Although, as in \cite{fedsr}, the scalers can be checked with random search, we propose for future work to learn them locally. Even with the same local architecture, each domain may overlap more or less with the others; thus, the local scalers should be adjusted dynamically.

Regarding the global scaler $\lambda_{fea}$, we observed that with higher values, the global model pulls the sources towards the target. Therefore, this effect impacts the local training at the client level while yielding larger source losses after communications. Together with the previous intuition from the local scalers, over several FL rounds, we could adapt the global one with a slow increment. This behavior will be tested in our future work.

\section{Source accuracy evaluation}
\subsection{FEDA - Federated Domain Generalisation experiments}
FEDA aims for generalisation on an unseen target domain. Despite this, we also overview the impact on the source local domains. For all three benchmarks evaluated, we confirm a fairly larger performance in average source accuracy when compared to the target domain. This shows that the model can generalise well to all the clients where the data is inferred during the FL training. FEDA achieves, on PACS, the closest performance source accuracy to the target at approximately 8\%.
\begin{table}[]
\centering
\resizebox{.47\textwidth}{!}{
\begin{tabular}{l|rcrcrclclc}
Dataset       & \multicolumn{10}{c}{FEDA - Domain/ Source Accuracy}                                                                                                                                                                             \\ \hline
PACS \cite{pacs}         & \cellcolor[HTML]{FFFFFF}A & \multicolumn{1}{c|}{\cellcolor[HTML]{FFFFFF}94.96} & \cellcolor[HTML]{FFFFFF}C & \multicolumn{1}{c|}{\cellcolor[HTML]{FFFFFF}92.67} & \cellcolor[HTML]{FFFFFF}P & \multicolumn{1}{c|}{93.69} & S & \multicolumn{1}{c|}{96.66} & Avg & 94.49 \\ \hline
OfficeHome \cite{officehome}   & \cellcolor[HTML]{FFFFFF}A & \multicolumn{1}{c|}{\cellcolor[HTML]{FFFFFF}91.2}  & \cellcolor[HTML]{FFFFFF}C & \multicolumn{1}{c|}{\cellcolor[HTML]{FFFFFF}80.21} & \cellcolor[HTML]{FFFFFF}P & \multicolumn{1}{c|}{84.4}  & R & \multicolumn{1}{c|}{86.26} & Avg & 85.52 \\ \hline
OfficeCaltech \cite{officecaltech} & \cellcolor[HTML]{FFFFFF}A & \multicolumn{1}{c|}{\cellcolor[HTML]{FFFFFF}75.87} & \cellcolor[HTML]{FFFFFF}C & \multicolumn{1}{c|}{\cellcolor[HTML]{FFFFFF}87.1}  & \cellcolor[HTML]{FFFFFF}D & \multicolumn{1}{c|}{80.68} & W & \multicolumn{1}{c|}{81.55} & Avg & 81.30
\end{tabular}
}
 \caption{FEDA - Federated Domain Generalisation \textbf{source} performance [average over 3 trials]}
  \label{tab:feda_sources}
\end{table}

\subsection{FEDALV - Federated Active Learning with Domain Generalisation experiments}
Before delving into the quantitative FEDALV evaluation of sources, it is necessary to mention the target accuracies of FEDA on OfficeCaltech when trained with domain per client and ResNet-18 architecture. Table \ref{tab:feda_officecaltech} displays the metrics according to the FDG settings of FPL \cite{flproto}, while the evaluation of FEDALV follows the configurations on PACS. Therefore, for each domain taken as a target, FEDA obtains the following accuracies: Amazon 90.45, Caltech 61.51, DSLR 77.08, Webcam 78.335, and an average of 76.84. According to Table \ref{tab:feda_officecaltech}, FEDALV attains a close target average performance of 72.34 with 50\% of the entire source dataset.

\begin{table}[]
\centering
\resizebox{.47\textwidth}{!}{
\begin{tabular}{l|c|c|c|c|c}
\begin{tabular}[c]{@{}l@{}}FAL Method / \\ \% of labelled data\end{tabular} & \multicolumn{1}{c|}{1\%}     & \multicolumn{1}{c|}{2\%}     & \multicolumn{1}{c|}{3\%}     & \multicolumn{1}{c|}{4\%}     & \multicolumn{1}{c}{5\%}      \\ \hline
Random                                                                      & \cellcolor[HTML]{FFFFFF}59.07 & \cellcolor[HTML]{FFFFFF}72.09 & \cellcolor[HTML]{FFFFFF}74.8  & \cellcolor[HTML]{FFFFFF}78.88 & \cellcolor[HTML]{FFFFFF}80.62 \\
CoreSet \cite{Sener2017ActiveApproach}                                            & \cellcolor[HTML]{FFFFFF}61.75 & \cellcolor[HTML]{FFFFFF}73.24 & \cellcolor[HTML]{FFFFFF}76.66 & \cellcolor[HTML]{FFFFFF}79.4  & \cellcolor[HTML]{FFFFFF}80.82 \\
LoGo \cite{logo}                                           & \cellcolor[HTML]{FFFFFF}61.95 & \cellcolor[HTML]{FFFFFF}73.39 & \cellcolor[HTML]{FFFFFF}77.45 & \cellcolor[HTML]{FFFFFF}79.57 & \cellcolor[HTML]{FFFFFF}81.13 \\
EADA \cite{eada}                                           & \cellcolor[HTML]{FFFFFF}59.81 & \cellcolor[HTML]{FFFFFF}71.89 & \cellcolor[HTML]{FFFFFF}77.74 & \cellcolor[HTML]{FFFFFF}82.24 & \cellcolor[HTML]{FFFFFF} \underline{84.63} \\
FEDAL (ours)                                                                & \cellcolor[HTML]{FFFFFF}62.56 & \cellcolor[HTML]{FFFFFF}69.21 & \cellcolor[HTML]{FFFFFF}74.61 & \cellcolor[HTML]{FFFFFF}78.16 & \cellcolor[HTML]{FFFFFF}79.43 \\
\rowcolor[HTML]{FFFFFF} 
\cellcolor[HTML]{FFFFFF}\textbf{FEDALV (ours)}                              & 62.5                          & 70.04                         & 76.56                         & 79.01                         & 79.92                        
\end{tabular}
}
 \caption{FEDALV - \textbf{PACS} dataset - AL \textbf{source} performance [average over 3 trials]}
  \label{tab:fedalv_pacs_source}
\end{table}

\begin{table}[]
\centering
\resizebox{.47\textwidth}{!}{
\begin{tabular}{l|c|c|c|c|c}
\begin{tabular}[c]{@{}l@{}}FAL Method / \\ \% of labelled data\end{tabular} & \multicolumn{1}{c|}{10\%}     & \multicolumn{1}{c|}{20\%}     & \multicolumn{1}{c|}{30\%}     & \multicolumn{1}{c|}{40\%}     & \multicolumn{1}{c}{50\%}      \\ \hline
Random                                                                      & \cellcolor[HTML]{FFFFFF}62.56 & \cellcolor[HTML]{FFFFFF}69.21 & \cellcolor[HTML]{FFFFFF}74.61 & \cellcolor[HTML]{FFFFFF}78.16 & \cellcolor[HTML]{FFFFFF}79.43 \\
CoreSet       \cite{Sener2017ActiveApproach}                                                                & \cellcolor[HTML]{FFFFFF}37.83 & \cellcolor[HTML]{FFFFFF}45.7  & \cellcolor[HTML]{FFFFFF}56.46 & \cellcolor[HTML]{FFFFFF}65.61 & \cellcolor[HTML]{FFFFFF}73.31 \\
LoGo       \cite{logo}                                                                    & \cellcolor[HTML]{FFFFFF}35.29 & \cellcolor[HTML]{FFFFFF}49.43 & \cellcolor[HTML]{FFFFFF}53.89 & \cellcolor[HTML]{FFFFFF}70.67 & \cellcolor[HTML]{FFFFFF}77.29 \\
EADA        \cite{eada}                                                                  & \cellcolor[HTML]{FFFFFF}37.75 & \cellcolor[HTML]{FFFFFF}53.21 & \cellcolor[HTML]{FFFFFF}58.8  & \cellcolor[HTML]{FFFFFF}70.74 & \cellcolor[HTML]{FFFFFF}78.69 \\
FEDAL (ours)                                                                & \cellcolor[HTML]{FFFFFF}36.08 & \cellcolor[HTML]{FFFFFF}52.94 & \cellcolor[HTML]{FFFFFF}59.61 & \cellcolor[HTML]{FFFFFF}71.16 & \cellcolor[HTML]{FFFFFF}77.76 \\
\rowcolor[HTML]{FFFFFF} 
\textbf{FEDALV (ours)}                                                      & 36.86                         & 61.66                         & 73.59                         & 74.69                         & \textbf{80.11}               
\end{tabular}
}
 \caption{FEDALV - \textbf{OfficeCaltech} dataset - AL \textbf{source} performance [average over 3 trials]}
  \label{tab:fedalv_officecaltech_source}
\end{table}

\begin{table}[]
\centering
\resizebox{.47\textwidth}{!}{
\begin{tabular}{l|c|c|c|c|c}
\begin{tabular}[c]{@{}l@{}}FAL Method / \\ \% of labelled data\end{tabular} & \multicolumn{1}{c|}{1\%}      & \multicolumn{1}{c|}{2\%}      & \multicolumn{1}{c|}{3\%}      & \multicolumn{1}{c|}{4\%}      & \multicolumn{1}{c}{5\%}                \\ \hline
Random                                                                      & \cellcolor[HTML]{FFFFFF}27.98 & \cellcolor[HTML]{FFFFFF}37.63 & \cellcolor[HTML]{FFFFFF}42.92 & \cellcolor[HTML]{FFFFFF}47.65 & \cellcolor[HTML]{FFFFFF}50.99          \\
CoreSet            \cite{Sener2017ActiveApproach}                                                              & \cellcolor[HTML]{FFFFFF}27.78 & \cellcolor[HTML]{FFFFFF}38.52 & \cellcolor[HTML]{FFFFFF}44.49 & \cellcolor[HTML]{FFFFFF}50.95 & \cellcolor[HTML]{FFFFFF}\underline{53.58} \\
LoGo          \cite{logo}                                                                & \cellcolor[HTML]{FFFFFF}28.2  & \cellcolor[HTML]{FFFFFF}36.75 & \cellcolor[HTML]{FFFFFF}40.96 & \cellcolor[HTML]{FFFFFF}46.8  & \cellcolor[HTML]{FFFFFF}50.63          \\
EADA                 \cite{eada}                                                     & \cellcolor[HTML]{FFFFFF}28.39 & \cellcolor[HTML]{FFFFFF}38.31 & \cellcolor[HTML]{FFFFFF}43.42 & \cellcolor[HTML]{FFFFFF}49.23 & \cellcolor[HTML]{FFFFFF}52.92          \\
FEDAL (ours)                                                                & \cellcolor[HTML]{FFFFFF}28.48 & \cellcolor[HTML]{FFFFFF}33.53 & \cellcolor[HTML]{FFFFFF}39.07 & \cellcolor[HTML]{FFFFFF}44.81 & \cellcolor[HTML]{FFFFFF}47.35          \\
\rowcolor[HTML]{FFFFFF} 
\textbf{FEDALV (ours)}                                                      & 28.12                         & 36.38                         & 43.49                         & 48.81                         & 53.48                                 
\end{tabular}
}
 \caption{FEDALV - \textbf{OfficeHome} dataset - AL \textbf{source} performance [average over 3 trials]}
  \label{tab:fedalv_officehome_source}
\end{table}

\begin{figure*}[!htp]
  \centering
 \includegraphics[trim=1cm .2cm 1cm 0.2cm, clip, width=.9\textwidth]{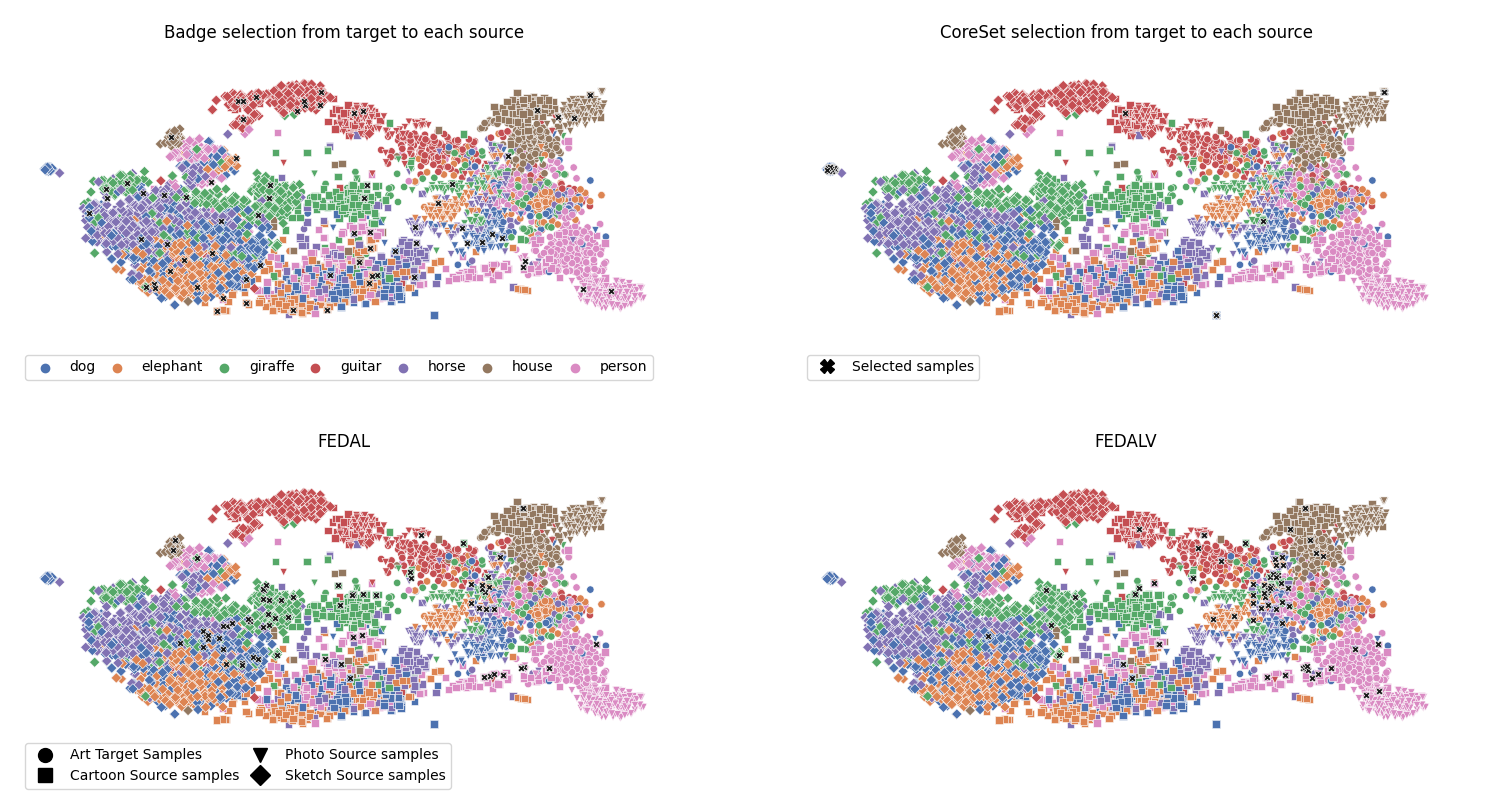}
  \caption{FEDALV: Different Selection functions [Zoom in for a better view]}
  \label{fig:fedalv_qual_supp}
\end{figure*}

Shifting the performance measurements of FEDALV on the source datasets, we gathered the results in Tables \ref{tab:fedalv_pacs_source}, \ref{tab:fedalv_officecaltech_source}, and \ref{tab:fedalv_officehome_source} for the three tackled datasets. We can observe that FEDALV yields competitive results against the other active learning (AL) baselines, especially for the OfficeCaltech dataset. The lack of consistent AL performance on source domains can be attributed to the uneven favouring of some clients over others during sampling to improve the target domain. Nevertheless, the decrease in performance happens at low budgets (1\%) and can be adjusted depending on the objective.

\section{FEDALV - Extensive Selection Analysis}
In our analysis of the selection function, we investigated several approaches for sampling informative data for the target domain. Additionally, we identified, as shown in Figure \ref{fig:fedalv_qual}, that target samples have higher free energies. To simplify our selection, we consider a fixed number of target samples to compare against the joint source distribution.

Therefore, given the target samples as centers, we deploy three metrics for selection: Badge \cite{badge}, CoreSet \cite{Sener2017ActiveApproach}, and the minimum Euclidean distance from FEDAL. Once selected within the defined budget, we calculate the Earth Moving Distance (EMD) between these new samples and the target ones. As expected, the source representations selected by FEDAL obtained the lowest EMD at 9.51 (compared to CoreSet 18.05 and 13.34 for Badge). 

In Figure \ref{fig:fedalv_qual_supp}, we marked with a cross the selected samples for each criterion after the first FAL cycle. The CoreSet methodology groups the furthest samples in regard to the target, while Badge selection has an even distribution over all the sources. However, both methods prove sub-optimal selection when aiming to reduce the misalignment with the target domain. On the other hand, FEDALV not only groups the selection closer to the high-energy target samples, but also where the classes are poorly clustered by the global model.